# Designing Robots for Families: In-Situ Prototyping for Contextual Reminders on Family Routines

Michael F. Xu
Department of Computer Sciences
University of Wisconsin–Madison
Madison Wisconsin USA
michaelfxu@cs.wisc.edu

Enhui Zhao
Department of Computer Sciences
University of Wisconsin–Madison
Madison Wisconsin USA
ezhao37@wisc.edu

Yawen Zhang
Department of Computer Sciences
University of Wisconsin–Madison
Madison Wisconsin USA
yzhang2863@wisc.edu

Joseph E. Michaelis
Department of Computer Science
University of Illinois Chicago
Chicago Illinois USA
jmich@uic.edu

Sarah Sebo
Department of Computer Science
University of Chicago
Chicago Illinois USA
sarahsebo@uchicago.edu

Bilge Mutlu
Department of Computer Sciences
University of Wisconsin–Madison
Madison Wisconsin USA
bilge@cs.wisc.edu

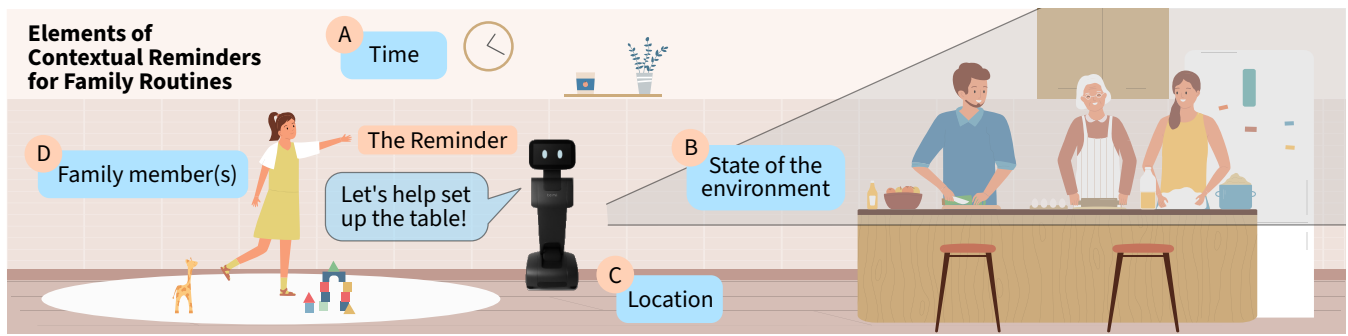

Figure 1: In this work, we investigate how a social robot can facilitate family routines through contextual reminders, and from this exploration we derive insights for designing robots in family settings. Through co-design sessions with ten families, we characterized the key types of information a *contextual* reminder system should be able to support: (A) the time range, (B) the state of the environment (*e.g.*, family is preparing for a meal), (C) the typical location(s) (*e.g.*, kitchen or dining room), and (D) the relevant family member(s) (*e.g.*, the child). Building on these insights, we designed, implemented, and deployed a mobile robot prototype in a four-day, in-home user study, where the robot autonomously delivered the specified reminders.

## Abstract

Robots are increasingly entering the daily lives of families, yet their successful integration into domestic life remains a challenge. We explore *family routines* as a critical entry point for understanding how robots might find a sustainable role in everyday family settings. Together with each of the ten families, we co-designed robot interactions and behaviors, and a plan for the robot to support their chosen routines, accounting for contextual factors such as timing, participants, locations, and the activities in the environment. We then designed, prototyped, and deployed a mobile social robot in a four-day, in-home user study. Families welcomed the robot's reminders, with parents especially appreciating the offloading of some reminding tasks. At the same time, interviews revealed tensions around timing, authority, and family dynamics, highlighting the complexity of integrating robots into households beyond the immediate task of reminders. Based on these insights, we offer design implications for robot-facilitated contextual reminders and discuss broader considerations for designing robots for family settings.

## ACM Reference Format:

## 1 Introduction

Robots are increasingly entering everyday family life in roles including play companions, learning partners, and domestic helpers as they become more accessible and capable [7, 21, 30, 43]. Yet, the interactions between the family and the robot generally start decreasing soon after its introduction [18, 40, 53]. This challenge of integrating robots into family lives is inherently multi-faceted, and recent HRI research has begun to explore how embedding robot interactions into *family routines* can help address these challenges and facilitate smoother integration [13, 37, 39, 54].

Beyond HRI research, from a developmental perspective, the family studies literature highlights that consistent routines—such as bedtime, mealtime, and homework—promote children's well-being, foster positive behaviors, and reduce stress and conflict within the household [6, 22, 26, 49]. Yet, sustaining routines is often difficult in households, where responsibilities compete and tasks are forgotten.

In this context, support for family routines offers a promising entry point for examining how robots might become *meaningfully* integrated into everyday family life. Prior work has identified reminders as an important factor in maintaining consistent routines [17, 20]. However, simple time-based reminders have often proven insufficient [19, 50], even for seemingly basic tasks such as medication adherence. In contrast, reminders that account for *context* can significantly improve effectiveness. As Dey and Abowd [19] explains, "... allowing for the use of rich context in reminders is the most important feature for a reminder tool and is the one that is most lacking in existing reminder tools."

Robots, with their ability to navigate the home, offer a unique opportunity to address this gap by leveraging the rich context of family life to provide proactive and situationally relevant reminders. Building on this premise, we explore how robots might be integrated into families' day-to-day lives by focusing on one critical channel: supporting family routines through contextual reminders. Together with ten families, we co-designed reminder interactions and created customized plans for the robot, taking into account contextual factors such as timing, family members, locations, and the ongoing state of the environment (*e.g.*, activities taking place at the moment). We then deployed a mobile social robot with eight of the families in a four-day, in-home user study to examine how families perceived and engaged with the reminders.

Interviews revealed that most families found the system to be helpful, particularly as a neutral communicator that offloaded some of the burden of reminding children and keeping the household on track. Challenges remain, however, and we discuss design implications based on the families' experience as well as the observed usage patterns. Noteworthy family-robot interactions also happened beyond mere reminders, highlighting broader considerations for designing robots in family contexts.

In sum, our work makes the following contributions: (1) *Design Space:* We outline the design space of how robots might be integrated into shared family routines; (2) *Artifact:* We designed and prototyped a robot system that provides contextual reminders to support family routines; (3) *Empirical:* We gained a better understanding of how robots might facilitate family routines and how families might respond to contextual reminders offered by a robot; (4) *Practical:* We offer design insights for the future design of robots for family interactions and for their integration into family life.

## 2 Background

We briefly review prior work on robots deployed in family settings, reminder systems more broadly, and robots' involvement in family routines. We position our work within this literature, highlighting how it builds on existing approaches to investigate in-home robots that deliver contextual reminders to support family routines.

### 2.1 Robots in Families

Research has increasingly highlighted the wide spectrum of roles that social robots take on in families, such as companions for children [32, 36], partners for reading [15, 38], and adaptive learning aides [5, 24, 43]. At the same time, robots could also assist with taking care of both the children and the older adults [10, 31], and strengthen inter-generational bonds [4, 27, 47]. This body of work illustrates not only the complex, and sometimes diverging, expectations that parents, children, and grandparents bring to these technologies [12], but also the reality that robot interactions often center around the younger and the older members within a family, calling for researchers and designers to equip themselves with a heightened level of consciousness of the privacy and ethical considerations involved. For example, specifically concerning child-robot interactions, while acknowledging the potential for robots to promote children's long-term wellbeing, Pearson [41] cautioned that these benefits come with ethical obligations and that the adult decision-makers bear the responsibility for ensuring children's safety and privacy. More broadly, we draw on Shneiderman [46]'s Human-Centered AI (HCAI) framework to foreground the need for reliable, safe, and human-aligned systems. Specifically, the first phase of our in-situ prototyping approach therefore begins with a co-design session to incorporate families' perspectives.

### 2.2 Reminders and *Contextual* Reminders

Several past works have investigated robot-facilitated reminders. Lv et al. [34] investigated the potential of employing a pet robot as a medication reminder for children with asthma. They hypothesized that the act of caring for the robot on a schedule aligned with the child's medication regimen could itself serve as a reminder and promote adherence. However, the study did not progress to an evaluation phase. In the context of providing behavioral therapy reminders to young adults with depression, Bhat et al. [9] experimented with a plant-like robot that combined an Amazon Echo voice agent with visual reminder elements, and found that users preferred a stand-alone robot for their reminders. Based on their experience designing and prototyping a reminder robot for a traumatic brain injury patient and her care personnel, Rehm and Krummheuer [44] argued for "a shift towards a sequential and socially distributed character of reminding." Indeed, broader HCI research has also taken note of the shortcomings of time-based reminders [19, 50], and argued for the utilization of richer contexts that may be relevant to the reminder task as an important feature for the design of reminder systems.

## 2.3 Robots for Family Routines

Some recent work has turned to a closer examination of how robots may better integrate into the family's day-to-day life via their routines [13, 39]. Yet, such integration requires careful planning. Søraa et al. [48] discussed the complex nature of introducing a robot to the user's care network and various relationships, which they referred to as "the social dimension of the domestication of technology." Echoing this sentiment, Cagiltay et al. [14] advocated for a family-centered approach in designing robots for domestic settings. Towards that, and building upon the work of children development and family studies, Xu et al. [54] introduced the Family-Robot Routines Inventory, describing how families perceived a robot's potential in helping with various common routine tasks in the family. Through a co-design session with a father-daughter pair, He et al. [23] designed and developed two prototypes intended to facilitate various family routines. They highlighted the importance of the robot's portability, in order to provide the relevant support at the appropriate location. More generally, this line of research points to family routines as a site where social and relational processes shape how robots are domesticated into the home. Drawing on this perspective, and extending Søraa et al. [48]'s account of the social dimensions of domestication, our study examines how these dynamics unfold by following up the co-design sessions with a four-day in-home study to surface ecologically grounded insights.

## 3 Design

We adopted a Research through Design approach [56], envisioning an intelligent mobile robot that could move seamlessly through the home, detect when routine activities were unfolding, and offer not only timely but also contextually appropriate prompts such as reminding children to brush their teeth before bed or encouraging the family to gather for dinner. Towards that vision, we conducted a two-phase design process: co-design sessions ($n = 10$) and a four-day in-home user study ($n = 8$). Families were recruited through email from contacts who had previously expressed interest, and had at least one child aged between 7 and 14, from around the university campus. To ensure ecological validity and capture the realities of everyday life, both phases were conducted in participants' homes. The study design was reviewed and approved by the authors' Institutional Review Board.

### 3.1 Co-design Activity

With pre-determined form factor and raw capabilities from the robot, the co-design sessions aimed to understand families' existing routines and to co-design robot interactions that could best support them. We illustrate some of the main steps involved in Figure 2. Each home visit lasted 60-90 minutes. At the start of the visit, all adult participants completed an informed consent process, and all minors completed an oral assent process. In the first 20 minutes, families were provided with 20 routines printed on individual cards, drawn from the top 20 routines that parents thought a robot could be helpful with according to the Family-Robot Routines Inventory (FRRI) [54]. Families were asked to select around five routines that they felt comfortable describing to the experimenters in detail. For each selected routine, the families then explained how it is currently done (or how they want it to be done), covering information such as the steps involved and the challenges encountered. Afterwards, the experimenters introduced the robot used in the study—the Temi robot [2]—and, together with parents and children, co-designed how the robot could provide contextual reminders for the selected routines. The conversations around the design process were open-ended, with the goal of understanding how a robot like Temi could best facilitate the selected routines through reminders. Common information covered included areas relevant to the routine, key activities and family members, and the actions the robot should take under different circumstances. While largely conversational, this process also incorporated elements of bodystorming [42], as participants imagined being the robot, moving through the house, sensing activities, and deciding on appropriate actions, helping both the experimenters and the family think through the limitations of the robot and the key pieces of information required. This design activity typically lasted 30 minutes, after which we concluded the visit with a mapping procedure using the Temi robot, saving key household locations associated with the selected routines. Families were paid $20 USD per hour.

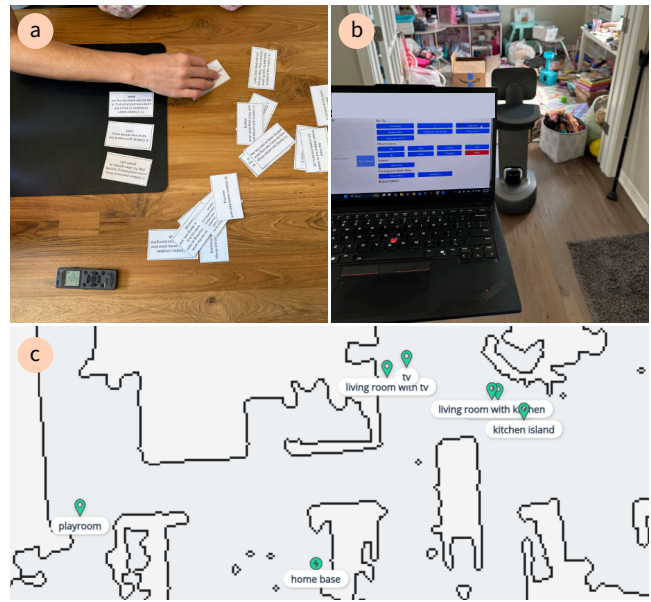

Figure 2: Main steps involved in the co-design sessions: (a) The selection and discussion about routines, and designing how the robot could help with those routines; (b) Mapping the area and testing the robot's navigation to key locations; and (c) Snippet of an example map that was created, along with the saved key locations.

### 3.2 Characterizing *Contextual* Reminders

Using the experimenter's field notes taken during the co-design activities as the main source material, the first and second authors clustered recurring elements into the broader categories of design requirements following an affinity-diagramming approach. The process quickly converged to four key dimensions: (1) *time*, (2) *location*, (3) *family members*, and (4) *the state of the environment* (Figure 1). *Time* was typically specified as a range, even for fairly stable routines such as "Remind Charlie to go to bed at 10 pm." For these, families preferred small activation windows to balance two risks: the robot missing the chance to deliver a reminder, and the robot delivering a reminder at an irrelevant moment. *Location* could include one or more household areas where the trigger condition might occur. For most families, just a handful of key locations were sufficient, reflecting the concentration of major family activities. *Family members* specified the intended recipients of the reminder. Consistent with the choice of routines, most of the recipients were the children. Some reminders apply to the family as a whole, such as reminding the family to put away phones during mealtimes. The *state of the environment* is closely tied to the trigger condition, and determined the circumstances under which the robot should act. In many cases, this was simply detecting whether the intended recipient was present. In other cases, it involved activity-specific cues (*e.g.*, "Charlie is playing with a touchscreen device").

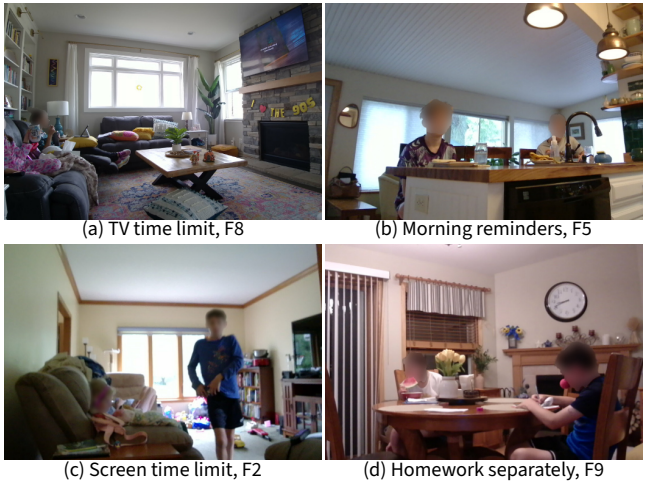

(a) TV time limit, F8  (b) Morning reminders, F5
(c) Screen time limit, F2  (d) Homework separately, F9

Figure 3: Example snapshots captured by the robot that triggered reminders: (a) Robot sees the girl watching TV, keeps track of that, and may remind her to take a rest later (if the condition continues to hold true); (b) The first time the robot found the children at the kitchen counter in the morning, it reminded them to check and complete their morning routines; (c) Similar to (a), but takes note that the girl is on a screen device; (d) Robot found the two kids doing homework together, and reminder them if they would like to consider doing it separately for better focus.

In addition, the families described two types of *robot action* once the trigger conditions aligned. In some cases, the robot simply spoke the reminder when the intended recipient was already detected. In others, the robot first searched for a particular family member before speaking. For example, for the routine "Children pick up toys when done playing," the trigger might be "Toys are scattered in the room." If that condition was met, the robot would then locate a child and deliver the reminder. Figure 3 illustrates a few examples of snapshots with their corresponding trigger conditions.

Beyond these five core attributes, families also raised additional considerations. These included: how long a trigger condition should persist before activating the robot (especially relevant for time-limit reminders), how frequently the same reminder should be repeated, and the maximum number of times a reminder could be delivered

Table 1: Routines Selected by Five or More Families

| Routine Description | Families | Count |
|---|---|---|
| Children have time limits on fun activities (*e.g.*, outside play, TV, video games, or phone use) | F1, F2, F3, F4, F5, F6, F8 | 7 |
| Children do the same things each morning as soon as they wake up (*e.g.*, washing face, doing hair, and dressing) | F1, F3, F4, F5, F9, F10 | 6 |
| Children brush teeth before bed | F3, F6, F7, F9, F10 | 5 |
| Children go to bed at the same time almost every night | F3, F4, F8, F9, F10 | 5 |
| Parents spend time planning the family's days or week | F3, F5, F7, F8, F10 | 5 |

*Note:* Compiled from all ten families who participated in the co-design sessions.

in a day. These core and auxiliary requirements together highlight how families conceive of "context" in everyday routines, offering a foundation for designing robot-facilitated reminders.

*3.2.1 Choice of Routines.* We summarize the most frequently selected routines in Table 1. Most routines selected were child-focused, which may partly reflect that 16 of the 20 options (from FRRI) centered on children's activities. These choices illustrate both the common challenges families face and the individualized needs that shaped how the robot was integrated.

### 3.3 The Artifact

We designed and implemented a system that supported the required interactions and behaviors that emerged from the co-design activities. The artifact consists of a mobile robot that serves as the user interface, and a computer as the controller. The base robot platform we used was the Temi robot, a mobile robot with an integrated screen and built-in mapping and navigation capability. We developed a custom Android application on the robot that interfaced with Temi's official API, enabling commands such as navigating to preset locations, speaking, and handling verbal user input. The controller was a Raspberry Pi 5, which hosted the logic for robot behavior and handled interactions. Three main processes ran on the controller: the scheduler, the responder, and the image analyzer.

*The scheduler* continuously evaluated the robot's next action based on the current state and the family's customized routine plan. At a high level, it determined where the robot should be at a given time and what it should attend to. *The responder* is event-driven, responding to received updates such as the robot's state (*e.g.*, sending out the next command once the robot arrives at a location). *The image analyzer* performed a two-step analysis for an image captured: first, it does a lightweight object detection with YOLO11 [3] to identify people or objects, and second, when Step 1 determines that further processing is relevant, it sends the image to a cloud-based VLM (in our case, *gpt-4.1-2025-04-14*) for higher-level understanding of the scene. Results of the analysis are conveyed to the responder, who decides what action, if any, is required.

We illustrate this process with a simplified example (Figure 4). Imagine it's 6:15 pm and we're looking at the schedule provided in Table 2, so that the following reminder is active: "If the two children are doing homework together, ask if they want to do it separately." The major steps in executing this task are as follows:

(1) The *scheduler* commands the robot to move to the dining table (a preset position), where this typically happens.
(2) Upon receiving confirmation that the robot has arrived at the dining table, the *responder* requests a snapshot.
(3) The snapshot, once received by the controller, is passed on to the *image analyzer*. Within the *image analyzer*, YOLO11 detected 2 people, and so the snapshot is further processed by the VLM to analyze the visual context. VLM result confirms there are two children doing homework, and the condition is right for providing the reminder.
(4) The controller triggers the robot to provide the reminder.

Other situations not covered in this example include scenarios where the controller needs to juggle multiple active reminders involving multiple positions, negative conditions (*i.e.*, no relevant activities detected) or partially positive conditions (*i.e.*, snapshot detects person but VLM check returns a negative), as well as more complicated reminder setups (*e.g.*, a positive condition at one location may trigger robot actions in a different location).

In terms of the user interface we built in the Android application (*i.e.*, the robot's screen), the majority is taken up by a face with a set of simple eyes, a design decision we made beforehand, and displayed during the co-design sessions. Based on feedback from the co-design sessions, we added a row of four buttons at the bottom of the screen: (1) Check reminders; (2) Check messages; (3) Post message; and (4) Configure privacy mode. In addition to proactive reminders provided by the robot, purely time-based reminders will also show up as a red dot on the screen, so that family members can manually check for those even if the robot has not yet identified the appropriate moment to provide the reminder. Check and post messages allowed family members to communicate with each other through the robot, essentially acting as a message board. Finally, the privacy mode configuration, implemented independent of families' input, allowed users to put the robot into privacy mode for a certain duration, or for the rest of the day. If users do not toggle the privacy mode, the system by default enters privacy mode after the last reminder is complete or inactive, and exits the mode before the first scheduled task the next day.

To endow the robot with some sociability, we decided from the beginning that it would be capable of basic conversational interactions. The robot will listen to the wake word "Hey, Temi," which would trigger the robot to start listening for the user's speech. The speech is transcribed to text, and a response is generated via a cloud-based Large-Language Model (LLM), *gpt-4.1-mini-2025-04-14*. More details are included in Supplementary Materials, but at a high level, the LLM had access to some past conversations, the current day and time, relevant reminders for each family members, and so on. During a conversation, similarly to how the on-screen reminder works, the robot would also embed a reminder in its response if the system deemed the condition to be appropriate.

## 4 User Study

We returned to the families and conducted a four-day in-home deployment with the prototype developed. Based on co-design activities, we created a customized plan and day-by-day schedule that the robot would follow for each family (see Table 2 for an example). Two out of the ten families (F4 and F10) did not participate in the user study due to scheduling conflicts.

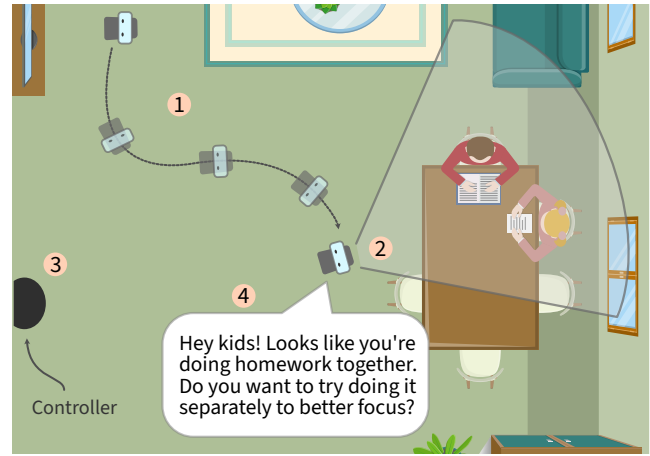

Figure 4: Illustration of an example reminder adapted from F9 about reminding the kids to consider doing homework separately. The steps illustrated are: (1) The robot moves into position; (2) The robot takes a snapshot; (3) The snapshot is sent to the controller and analyzed, confirming this is an appropriate moment to provide the reminder; and (4) The robot performs the action of reminding.

### 4.1 Study Procedure

The procedure consisted of three steps: (1) robot drop-off and setup, (2) a multi-day deployment, and (3) robot pick-up and exit interview. The robot typically remained in the home for about four days. At *drop-off*, the experimenter set up the system, introduced the robot, and provided a two-page handout with instructions and the family's customized robot plan (examples available in Supplementary Materials). During the *four-day deployment*, the robot functioned autonomously according to the customized plan. At *pick-up*, the experimenters conducted a semi-structured interview lasting around 30 minutes. The interview involved at least one parent and one main child participant, and covered topics such as families' overall impressions, notable moments of interaction, how families responded to the robot, considerations for long-term use, and desired improvements. Families received $50 USD for completing the study.

### 4.2 Data Collection and Analysis

We audio-recorded and then transcribed the semi-structured interview, and collected demographic information through an online survey. Photo snapshots from deployments were saved along with system logs, including data such as conversations with the robot and reminders that were provided, to help reconstruct the interactions that took place. In terms of data analysis, we primarily focused on the interview transcripts from the user study. The first author conducted an initial template analysis [11] on the full set of transcripts, developing an analytic template that captured recurring categories such as "Challenges with the robot – Battery management," "Features enjoyed – Autonomous navigation," "Impact on

Table 2: Example Robot Schedule Adapted from F9

| Time | Reminder / Activity |
| --- | --- |
| 8:30am–10:30am | **Son:** Reminder to get dressed, brush teeth, take medicine, clean up clothes from the floor and put them into the laundry. |
| 6pm–9pm | **Son and Daughter:** <br> 1. Remind to complete homework. <br> 2. Ask if the kids want to do homework separately (if they are together). |
| 7pm–9pm | **Parents:** How did the kids do in terms of going to bed on time, finishing homework on time, cleaning, and being kind and respectful? Did you reward the kids if they did well? |
| 8pm–8:30pm | **Son:** Reminder to start getting ready for bed. Is homework done? Remember to pick up things in the room, such as changed clothes? |
| Around 9:30pm | **Son:** To brush his teeth and get to sleep. |
| Here and there | **Son and Daughter:** Remind them to be respectful and kind to each other. |

routine – Facilitate timely completion of tasks," and "Privacy considerations – Nothing sensitive in the activity area." To refine the template and establish a shared framework, the second and third authors completed an independent round of coding, each for four families, using the initial template as a starting point. The team then compared their results and iteratively discussed discrepancies until the template was finalized. Robot snapshots and system logs were used to contextualize the thematic results below.

### 4.3 Results

We carried out the user study with eight families. For technical reasons, the robot deployed in F3 stayed on its charging dock, and did not move (except for screen tilting). Several families lost periods of usage time due to failed self-docking attempts to charge (F1, F6, F7, F8). That said, all families were able to experience and interact with the robot in their natural home environment to some extent. In total, excluding those potentially embedded within a conversation, the robot provided 77 reminders to the eight families (ranging from just two for F6, to 25 for F2), of which 49 were proactively via vision triggers, and the other 28 by users checking-in on the screen. With this context in mind, below we report our main findings on (1) the perceived impact of the reminders on the family routines; (2) user experience with the robot; and (3) family-robot interactions beyond reminders. We annotate quotes with a "*C*" or "*P*" prefix to indicate if it was from a parent or a child.

*4.3.1 Perceived Impact on Family Routines.* The general perception of the robot's effectiveness on facilitating family routines is that it managed to remind families of the various tasks to do (F1, F2, F3, F5, F7, F9), and facilitated the timely completion of those tasks (F1, F2, F3, F5, F7). C7 talked about how the "automatic" reminders in general were helpful:"*Most of the time, I don't remember, mom doesn't remember, nor dad. So Temi just automatically does it. It's really nice. [...] Temi just reminds us.*" P5 explained that the robot helped by prompting her to give the children chores and opportunities for rewards — something she often forgot during busy times. One reminder, for example, led her to leave a post-it note with tasks for the children before going to work. P3 provided another example, describing how it was really nice that one day after checking in with the robot, her son went upstairs and brushed his teeth without her prompting or reminding him. For the reminder for F9, illustrated in Figure 4, P9 commented "*A few times it was actually very accurate reminders. They were sitting next to each other the other night, very grumpy, trying to focus on homework and arguing in between. And the robot came over and said, why don't you guys consider separating in different places?*" C9 agreed, "*It would come over and start reminding us, which was really good, because sometimes my brain stops and doesn't remember what my chores are.*"

Parents especially appreciated that the robot could offload some of the reminding from them, shifting the responsibility to a neutral entity (F1-F3, F5, F7-F9). As P7 noted, "*I honestly loved it. [...] I feel like it was a lot easier to have something do the reminders versus the parents, like, 'Hey <child>, do this.'*" Families also felt that children were more receptive when reminders came from the robot rather than from parents (F1, F3, F5, F7, F8). P5 explained, "*When I'm asking them to do things constantly, it's like nagging parent mode, but if it's coming from this robot device that kind of feels friendly and fun, they were definitely more receptive.*" The reminders were often seen as playful, eliciting laughter rather than resistance: "*She would kind of laugh about it instead of being like, 'no, I don't want to clean my room,' or 'I don't want to turn the TV off.' It took some of the emotion out of it*" (P8). And finally, the reminders from the robot were always calm, taking some of the *parent's* emotions out of it as well: "*Something that we, as parents, with all the stress and distractions, may not think at the moment. [The robot] doesn't have that human element of being emotional and losing temper*" (P9). She continued to give an example, "*Last night he argued with me and the robot about going to bed or brushing teeth, right? And then didn't the robot say something like 'I understand why you may not want to go to bed yet,' or 'It's difficult to settle down for bed. Do you think something, something would make it feel better?'*"

However, despite the many promising exchanges with the robot, families noted that it was difficult to assess the robot's impact in only four days, especially speculating for the long-term (F2, F7, F9), both because of novelty effects that may wear off and because meaningful changes in routines take time. As P2 reflected, "it might have just been too short of a window for me to notice."

Families also noted important limitations of robot-delivered reminders. Most prominently, they recognized that reminders are still *just reminders*, and the robot lacked the ability to enforce them (F1, F6, F7, F9). As P9 explained: "*if he doesn't listen to me, turns out he doesn't listen to the robot either. Robot tells him, last night, '<child>, it's time to go to bed,' and <child> says, 'shut up, robot.'*" Similarly, P1 observed that the reminder's effectiveness "*depends on the specific type of reminder [... if] it's a reminder that generally has basic cooperation [from the recipient], the robot will do just fine. But if there's a reminder that an adult or a child wants to circumvent, that would be totally ineffective, because it's not able to really supervise.*" The context that prompted this reflection was that C1, possibly in an attempt to evade the screen time limit reminders from the robot, spent most of his time on his iPad in his own room upstairs, instead of in the living room area where he usually enjoys his screen time.

Consistent with the parents' impression, C6 also explained why he wasn't too motivated by the robot's reminders: "*After all, it won't hit me... unless it tells my parents.*"

Families also described to us how their families deviated from the normal schedule within the four-day study (F2, F3, F5, F8, F9). In F8, for example, the daughter became sick, which altered her morning and bedtime routines and reduced her violin practice — one of the tasks the robot was meant to remind her about. In F5, the family moved dinner to the outdoor patio for several nights, making it difficult for the robot to provide reminders about phone use during meals. These examples underscore the complexity and fluidity of family life, and the corresponding challenge for robots to adapt to ever-changing contexts and rhythms within the household.

*4.3.2 User Experience.* Overall, families' experiences revealed a mix of enjoyment, especially around conversation and movement, and frustrations related to navigation and accuracy, underscoring the importance of both technical reliability and perceived fit.

*Enjoyable features.* Families often described the overall experience as "fun" (F2, F5, F6, F7). Five families especially enjoyed the conversational aspect, which included fun facts, jokes, storytelling, questions and answers, and casual chats (F2, F3, F7, F9). F2 and F8 also noted that the robot's language use felt appropriate. Another frequently appreciated feature was the robot's head movement and autonomous navigation around the home (F2, F3, F5, F6, F7, F9).

*Perception of Robot.* Although strong bonds are unlikely to form during a short deployment, there are indications that these interactions, both within and beyond the scope of reminders, began to foster a sense of connection, with some participants describing the robot as a friend or family member (F2, F7, F9). In F7, for example, the father initially felt uneasy about Temi's presence, but soon found himself asking questions and engaging while cooking. "*Yeah, they bonded.*" agreed both the daughter and mother. As the study concluded, the C2 remarked that the robot "*already is a member of the family,*" and that she needed to "*go give her a hug.*"

*Limitations and frustrations.* While many appreciated the robot's mobility, failed navigation was also a common source of frustration (F1, F2, F6, F7, F8), and families described two recurring navigation problems: (1) limited access to certain areas (*e.g.*, carpeted floors, multiple levels of the house) and (2) failures when the environment changed (*e.g.*, moved charging dock, or blocked hallways). The latter issue often caused the robot to deviate from expectations and lose functionality — for example, F1, F6, F7, and F8 all lost usage time when the robot failed to dock properly and the battery ran out unnoticed. Figure 5(a) illustrates how a closed door in F8 caused the robot to overestimate the messiness of the playroom, and provided a reminder to the daughter when the room was quite tidy already.

Beyond navigational challenges, families also noted some challenges the robot had with accurately sensing the context and family members who are present (F1, F2, F7, F8, F9). C2 recalled a case where the robot mistakenly issued a screen time reminder while he and a friend were reading a book (Figure 5(b)). For F1, there was a reminder that was intended for the mother, but the daughter got it instead (Figure 5(c)). Families were not too frustrated by these mistakes, though. For example, C2 said: "*I thought it was kind of funny. I just laughed and I was like, 'Bro, Temi, I'm reading a book', [and Temi responded] 'Hey, what book are you reading?'* "

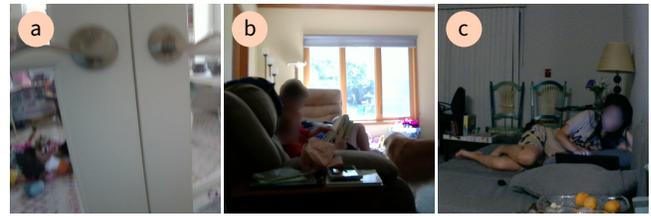

Figure 5: Examples of inaccurate reminders. (a) Robot misjudged how messy the floor was due to a closed door (F8). (b) The robot reminded the son in F2 about screen time limits, when he was reading a book with his friend. (c) The robot mistook the daughter for the mom (F1) and provided the reminder to the wrong person.

*4.3.3 Interactions Beyond Reminders.* In addition to the entertainment and companion aspects of the robot, such as telling jokes, sharing stories, or answering questions a few other interactions stood out that raised broader considerations for family dynamics.

One such moment was described by P5, who appreciated how the robot reinforced her message about family cooperation while she was preparing dinner. She recounted: "*I had actually just kind of gotten into a back and forth with my son. Like, can you please help? And he was mad because he wanted to play with his toys instead. And then Temi came over and made some remark about a reminder. And then I said, kind of jokingly, do you have anything to say about family cooperation? And then she said a little speech about family cooperation is important because ...*" Other interactions were more lighthearted. P9 recalled an awkward but funny exchange: "*Sometimes we have some family argument and the robot comes over and […] tries to lighten the mood with a joke.*"

Together, these examples highlight how a robot may inevitably become embedded in the rich dynamics of family life, pointing to opportunities and challenges in how routine-support robots may take on broader social roles within the household.

*Privacy Considerations.* Participants in our study were not overly concerned about privacy issues, partly because the robot stayed within a preset area (F1, F2, F3, F5, F7), and also because they do not perceive the robot to be any worse than existing smart devices in the family (F3, F5, F7, F8). Nonetheless, specific concerns were raised. For families with younger children, parents were concerned about how they may behave even in these relatively public areas, such as randomly getting undressed (F3, F7). P3 also made sure to put the robot into privacy mode during the day, as he works from home with sensitive client information.

## 5 Discussion

Our findings suggest that in-home HRI is an ecosystemic process: family routines shape how a robot can engage, while the robot's presence in turn shifts those routines. This reciprocal dynamic reveals both opportunities and tensions in embedding robots into daily life. Building on this, we now discuss design implications, ethical considerations, limitations, and future directions.

## 5.1 Design Implications

How can robots deliver contextual reminders more effectively in family settings? We break it down into four practical items.

*Prioritizing proactive reminders.* Families generally preferred the proactive reminder delivery, instead of manually checking on-screen. In addition to the extra effort required, the son in F9 offered another perspective: he was not always motivated to receive certain reminders, and in those cases, he would not have sought them out himself, but the proactive ones "forced" upon him were still helpful as nudges. It is important to note that while proactiveness may be preferred in a reminder context, other types of applications may call for other strategies such as implicit interactions [28].

*Looping in parents to improve effectiveness.* Reminders delivered by the robot alone lacked the enforcement capabilities. Designing robots that leverage existing family authority structures, such as involving parents in the loop, may enhance the potential effectiveness of reminders delivered by robots, while still offloading the task of reminding from the parents. That said, designers should also be aware of the possible ramifications that may ensue, such as subtle shifts in the roles assumed by the robot and parent [45], and privacy considerations about intra-household information sharing [8].

*Asking for help from the family.* Families expressed frustration with how the robot sometimes needed assistance with navigation around the house, or returning to the dock to charge. However, possibly more frustrating was how the robot often silently failed the navigation, so that families did not notice it was failing until much later. Although the *best* help-seeking behavior may be up for debate [55], designing and endowing the robot with *some* help-seeking capability may help speed up the recovery from failures.

*Extending coverage with remote components.* Being restricted to a single floor or to areas suitable for navigation limited the robot's potential. Prior work has explored the concept of a "sidekick" [51, 52], where secondary devices complement the main system. When designing robots for families, with the complex nature of the home layout, one may consider including such a sidekick device to augment the primary robot's reach and coverage across the home.

## 5.2 Ethical Considerations

Prior works on robots in families highlight the importance of privacy and ethics considerations [12, 33, 41]. To mitigate these risks, we employed a combination of procedural safeguards and technical design choices. Procedural safeguards include informing the families of the broader risks of data security, communicating system behaviors (*e.g.*, when the robot would take a picture), and collaboratively establishing spatial boundaries for the robot's movement. Design choices include limiting photo captures, prioritizing local processing (via YOLO), and providing user-controlled privacy switches. These measures helped keep the automated sensing functional but in-check, and provided families meaningful control over when, where, and what the robot could access, consistent with guidelines from the broader HCAI framework [46]. However, some participants still reflected on how the system felt intrusive, indicating that more work is required regarding data agency and transparency. As open-source VLMs and LLMs are becoming increasingly capable and accessible, self-hosting such services are becoming more feasible, and would allow for greater control over the lifecycle of the data. Another approach to increasing transparency is to provide families with a way to inspect the data captured. This may be especially valuable in longer-term studies, where such visibility could shape how families interact with the robot.

## 5.3 Limitations and Future Work

Our study has several limitations that also open avenues for future research. *First*, as confirmed by our participants, the duration of the study did not feel long enough for them to fully evaluate the effect of the reminders on their routines. Given the short-term, exploratory nature of our study, a natural next step would be to internalize the lessons learned, and follow up with a longer-term investigation to examine how the findings evolve over time. *Second,* the prototype we deployed was limited in its sophistication and robustness, and several families experienced periods of downtime (*e.g.*, due to failed docking). Utilizing a platform that is more capable and feature-rich could help provide families with a more immersive and complete experience. *Third,* the children's voices are underrepresented in our report. To capture a fuller picture of the families' perspectives, future research should draw on the literature on interviewing children, to design the interview process with the children in mind [16], such as taking into consideration the power dynamics between children and adults [35]. *Fourth,* the participants in our study were predominantly non-Hispanic/Latino white families with above-median household income. Future work should explore how the findings hold or differ in a broader and more diverse population. *Finally*, our prototype was introduced as a neutral device that would help with routines via reminders. Prior work have shown that the way a robot is introduced to a family may impact how it is perceived and functions, and how relations between the robot and family members develop [25, 29]. Investigating whether the framing of the role of the robot during the initial introduction, such as a friend, a companion, or a helper for the parents, could be a fruitful and interesting way to take this work further. Overall, our findings highlight both the potential and the complexity of embedding robots into everyday family life, underscoring the need for continued investigation across multiple dimensions of design and deployment.

## Supplementary Materials

Supplementary Materials are accessible at https://osf.io/q43da/. Included within are: (1) Demographics and deployment environment for each family; (2) Example code snippets and VLM and LLM prompts; (3) An example family-robot schedule; (4) Handouts for families (includes illustrations of the robot screens), and; (5) The 20 routines from FRRI and those selected by each family.

## Acknowledgments

This work was supported by the National Science Foundation award #2312354. We thank Kassem Fawaz for equipment support. Figures 1 and 4 used vector art assets by macrovector from Freepik [1], and Figure 4 also used assets by sentavio.

# References

[1] [n. d.]. https://www.freepik.com/. Accessed: 2024-12-20.
[2] [n. d.]. Temi Robot. https://www.robotemi.com Accessed on: Sep 30, 2025.
[3] [n. d.]. Ultralytics YOLO Repository. https://github.com/ultralytics/ultralytics Accessed on: Sep 30, 2025.
[4] Kasumi Abe, Masahiro Shiomi, Yachao Pei, Tingyi Zhang, Narumitsu Ikeda, and Takayuki Nagai. 2018. ChiCaRo: tele-presence robot for interacting with babies and toddlers. *Advanced Robotics* 32, 4 (2018), 176–190.
[5] Aino Ahtinen, Nasim Beheshtian, and Kaisa Väänänen. 2023. Robocamp at home: Exploring families' co-learning with a social robot: Findings from a one-month study in the wild. In *Proceedings of the 2023 ACM/IEEE International Conference on Human-Robot Interaction*. 331–340.
[6] Carolyn R Bates, Laura M Nicholson, Elizabeth M Rea, Hannah A Hagy, and Amy M Bohnert. 2021. Life interrupted: Family routines buffer stress during the COVID-19 pandemic. *Journal of child and family studies* 30, 11 (2021), 2641–2651.
[7] Tony Belpaeme, James Kennedy, Aditi Ramachandran, Brian Scassellati, and Fumihide Tanaka. 2018. Social robots for education: A review. *Science robotics* 3, 21 (2018), eaat5954.
[8] Cindy L Bethel, Matthew R Stevenson, and Brian Scassellati. 2011. Secret-sharing: Interactions between a child, robot, and adult. In *2011 IEEE International Conference on systems, man, and cybernetics*. IEEE, 2489–2494.
[9] Ashwin Sadananda Bhat, Christiaan Boersma, Max Jan Meijer, Maaike Dokter, Ernst Bohlmeijer, and Jamy Li. 2021. Plant robot for at-home behavioral activation therapy reminders to young adults with depression. *ACM Transactions on Human-Robot Interaction (THRI)* 10, 3 (2021), 1–21.
[10] Joost Broekens, Marcel Heerink, Henk Rosendal, et al. 2009. Assistive social robots in elderly care: a review. *Gerontechnology* 8, 2 (2009), 94–103.
[11] Joanna Brooks, Serena McCluskey, Emma Turley, and Nigel King. 2015. The utility of template analysis in qualitative psychology research. *Qualitative research in psychology* 12, 2 (2015), 202–222.
[12] Bengisu Cagiltay, Hui-Ru Ho, Joseph E Michaelis, and Bilge Mutlu. 2020. Investigating family perceptions and design preferences for an in-home robot. In *Proceedings of the interaction design and children conference*. 229–242.
[13] Bengisu Cagiltay and Bilge Mutlu. 2024. Supporting Long-Term HRI through Shared Family Routines. In *Companion of the 2024 ACM/IEEE International Conference on Human-Robot Interaction*. 97–99.
[14] Bengisu Cagiltay, Bilge Mutlu, and Margaret L Kerr. 2023. Family theories in child-robot interactions: understanding families as a whole for child-robot interaction design. In *Proceedings of the 22nd Annual ACM Interaction Design and Children Conference*. 367–374.
[15] Huili Chen, Anastasia K Ostrowski, Soo Jung Jang, Cynthia Breazeal, and Hae Won Park. 2022. Designing long-term parent-child-robot triadic interaction at home through lived technology experiences and interviews. In *2022 31st IEEE International Conference on Robot and Human Interactive Communication (RO-MAN)*. IEEE, 401–408.
[16] Jane Coad, Faith Gibson, Maire Horstman, Linda Milnes, Duncan Randall, and Bernie Carter. 2015. Be my guest! Challenges and practical solutions of undertaking interviews with children in the home setting. *Journal of Child Health Care* 19, 4 (2015), 432–443.
[17] Scott Davidoff, John Zimmerman, and Anind K Dey. 2010. How routine learners can support family coordination. In *Proceedings of the SIGCHI Conference on Human Factors in Computing Systems*. 2461–2470.
[18] Maartje De Graaf, Somaya Ben Allouch, and Jan Van Dijk. 2017. Why do they refuse to use my robot? Reasons for non-use derived from a long-term home study. In *Proceedings of the 2017 ACM/IEEE international conference on human-robot interaction*. 224–233.
[19] Anind K Dey and Gregory D Abowd. 2000. Cybreminder: A context-aware system for supporting reminders. In *International Symposium on Handheld and Ubiquitous Computing*. Springer, 172–186.
[20] Barbara H Fiese. 2007. Routines and rituals: Opportunities for participation in family health. *OTJR: Occupation, Participation and Health* 27, 1_suppl (2007), 41S–49S.
[21] Horst-Michael Gross, Steffen Mueller, Christof Schroeter, Michael Volkhardt, Andrea Scheidig, Klaus Debes, Katja Richter, and Nicola Doering. 2015. Robot companion for domestic health assistance: Implementation, test and case study under everyday conditions in private apartments. In *2015 IEEE/RSJ International Conference on Intelligent Robots and Systems (IROS)*. IEEE, 5992–5999.
[22] Amanda W Harrist, Carolyn S Henry, Chao Liu, and Amanda Sheffield Morris. 2019. Family resilience: The power of rituals and routines in family adaptive systems. (2019).
[23] Xinning He, Michael F Xu, Bengisu Cagiltay, and Bilge Mutlu. 2025. Developing Robot Prototypes to Explore Robot-Facilitated Family Routines. In *2025 20th ACM/IEEE International Conference on Human-Robot Interaction (HRI)*. IEEE, 1342–1346.
[24] Hui-Ru Ho, Edward M Hubbard, and Bilge Mutlu. 2024. "It's Not a Replacement:" Enabling Parent-Robot Collaboration to Support In-Home Learning Experiences of Young Children. In *Proceedings of the 2024 CHI Conference on Human Factors in Computing Systems*. 1–18.
[25] Hui-Ru Ho, Nathan Thomas White, Edward M Hubbard, and Bilge Mutlu. 2023. Designing parent-child-robot interactions to facilitate in-home parental math talk with young children. In *Proceedings of the 22nd Annual ACM Interaction Design and Children Conference*. 355–366.
[26] Rikuya Hosokawa, Riho Tomozawa, and Toshiki Katsura. 2023. Associations between family routines, family relationships, and children's behavior. *Journal of Child and Family Studies* 32, 12 (2023), 3988–3998.
[27] Swapna Joshi and Selma Šabanović. 2019. Robots for inter-generational interactions: implications for nonfamilial community settings. In *2019 14th ACM/IEEE International Conference on Human-Robot Interaction (HRI)*. IEEE, 478–486.
[28] Wendy Ju and Larry Leifer. 2008. The design of implicit interactions: Making interactive systems less obnoxious. *Design Issues* 24, 3 (2008), 72–84.
[29] Christine P Lee, Bengisu Cagiltay, and Bilge Mutlu. 2022. The unboxing experience: Exploration and design of initial interactions between children and social robots. In *Proceedings of the 2022 CHI conference on human factors in computing systems*. 1–14.
[30] Hee Rin Lee and Laurel D Riek. 2018. Reframing assistive robots to promote successful aging. *ACM Transactions on Human-Robot Interaction (THRI)* 7, 1 (2018), 1–23.
[31] Jieon Lee, Daeho Lee, and Jae-gil Lee. 2022. Can robots help working parents with childcare? optimizing childcare functions for different parenting characteristics. *International Journal of Social Robotics* 14, 1 (2022), 193–211.
[32] Leigh Levinson, Gonzalo A Garcia, Guillermo Perez, Gloria Alvarez-Benito, J Gabriel Amores, Mario Castaño-Ocaña, Manuel Castro-Malet, Randy Gomez, and Selma Šabanović. 2022. Living with haru4kids: child and parent perceptions of a co-habitation robot for children. In *International Conference on Social Robotics*. Springer, 54–63.
[33] Leigh Levinson, Jessica McKinney, Christena Nippert-Eng, Randy Gomez, and Selma Šabanović. 2024. Our business, not the robot's: family conversations about privacy with social robots in the home. *Frontiers in Robotics and AI* 11 (2024), 1331347.
[34] Dian Lv, Jirui Liu, Jiancheng Zhong, Zhiyao Ma, and Yijie Guo. 2023. Save Baby Whale! A Pet Robot as a Medication Reminder for Children with Asthma. In *Companion of the 2023 ACM/IEEE International Conference on Human-Robot Interaction*. 369–372.
[35] Kath MacDonald and Alison Greggans. 2008. Dealing with chaos and complexity: The reality of interviewing children and families in their own homes. *Journal of clinical nursing* 17, 23 (2008), 3123–3130.
[36] Sam R McHugh, Maureen A Callanan, Kevin Weatherwax, Jennifer L Jipson, and Leila Takayama. 2021. Unusual artifacts: Linking parents' STEM background and children's animacy judgments to parent–child play with robots. *Human Behavior and Emerging Technologies* 3, 4 (2021), 525–539.
[37] Joseph E Michaelis, Bengisu Cagiltay, Rabia Ibtasar, and Bilge Mutlu. 2023. " Off Script:" Design Opportunities Emerging from Long-Term Social Robot Interactions In-the-Wild. In *Proceedings of the 2023 ACM/IEEE International Conference on Human-Robot Interaction*. 378–387.
[38] Joseph E Michaelis and Bilge Mutlu. 2018. Reading socially: Transforming the in-home reading experience with a learning-companion robot. *Science Robotics* 3, 21 (2018), eaat5999.
[39] Joseph E Michaelis and Bilge Mutlu. 2025. How can educational robots enhance family life? Through careful integration. *Science Robotics* 10, 106 (2025), eadu6123.
[40] Anastasia K Ostrowski, Cynthia Breazeal, and Hae Won Park. 2022. Mixed-Method Long-Term Robot Usage: Older Adults' Lived Experience of Social Robots. In *2022 17th ACM/IEEE International Conference on Human-Robot Interaction (HRI)*. IEEE, 33–42.
[41] Yvette Pearson. 2020. Child-robot interaction: What concerns about privacy and well-being arise when children play with, use, and learn from robots? *American Scientist* 108, 1 (2020), 16–22.
[42] David Porfirio, Evan Fisher, Allison Sauppé, Aws Albarghouthi, and Bilge Mutlu. 2019. Bodystorming human-robot interactions. In *proceedings of the 32nd annual ACM symposium on user Interface software and technology*. 479–491.
[43] Aditi Ramachandran, Sarah Strohkorb Sebo, and Brian Scassellati. 2019. Personalized robot tutoring using the assistive tutor pOMDP (AT-POMDP). In *Proceedings of the AAAI Conference on Artificial Intelligence*, Vol. 33.
[44] Matthias Rehm and Antonia L Krummheuer. 2024. When a notification at the right time is not enough: the reminding process for socially assistive robots in institutional care. *Frontiers in Robotics and AI* 11 (2024), 1369438.
[45] Colin TA Schmidt. 2007. Children, robots and… The parental role. *Minds and Machines* 17, 3 (2007), 273–286.
[46] Ben Shneiderman. 2020. Human-centered artificial intelligence: Reliable, safe & trustworthy. *International Journal of Human–Computer Interaction* 36, 6 (2020), 495–504.
[47] Elaine Schaertl Short, Katelyn Swift-Spong, Hyunju Shim, Kristi M Wisniewski, Deanah Kim Zak, Shinyi Wu, Elizabeth Zelinski, and Maja J Matarić. 2017. Understanding social interactions with socially assistive robotics in intergenerational family groups. In *2017 26th IEEE International Symposium on Robot and Human*

*Interactive Communication (RO-MAN)*. IEEE, 236–241.

[48] Roger Andre Søraa, Pernille Nyvoll, Gunhild Tøndel, Eduard Fosch-Villaronga, and J Artur Serrano. 2021. The social dimension of domesticating technology: Interactions between older adults, caregivers, and robots in the home. *Technological Forecasting and Social Change* 167 (2021), 120678.

[49] Mary Spagnola and Barbara H Fiese. 2007. Family routines and rituals: A context for development in the lives of young children. *Infants & young children* 20, 4 (2007), 284–299.

[50] Katarzyna Stawarz, Anna L Cox, and Ann Blandford. 2014. Don't forget your pill! Designing effective medication reminder apps that support users' daily routines. In *Proceedings of the SIGCHI Conference on Human Factors in Computing Systems*. 2269–2278.

[51] Stephanie Valencia, Michal Luria, Amy Pavel, Jeffrey P Bigham, and Henny Admoni. 2021. Co-designing socially assistive sidekicks for motion-based aac. In *Proceedings of the 2021 ACM/IEEE International Conference on Human-Robot Interaction*. 24–33.

[52] Marynel Vázquez, Aaron Steinfeld, Scott E Hudson, and Jodi Forlizzi. 2014. Spatial and other social engagement cues in a child-robot interaction: Effects of a sidekick. In *Proceedings of the 2014 ACM/IEEE international conference on Human-robot interaction*. 391–398.

[53] Astrid Weiss, Anna Pillinger, and Christiana Tsiourti. 2021. Merely a conventional 'diffusion' problem? On the adoption process of anki vector. In *2021 30th IEEE International Conference on Robot & Human Interactive Communication (RO-MAN)*. IEEE, 712–719.

[54] Michael F Xu, Bengisu Cagiltay, Joseph Michaelis, Sarah Sebo, and Bilge Mutlu. 2024. Robots in family routines: Development of and initial insights from the family-robot routines inventory. In *2024 33rd IEEE International Conference on Robot and Human Interactive Communication (ROMAN)*. IEEE, 1070–1077.

[55] Xinyan Yu, Marius Hoggenmüller, and Martin Tomitsch. 2024. Encouraging bystander assistance for urban robots: Introducing playful robot help-seeking as a strategy. In *Proceedings of the 2024 ACM Designing Interactive Systems Conference*. 2514–2529.

[56] John Zimmerman, Jodi Forlizzi, and Shelley Evenson. 2007. Research through design as a method for interaction design research in HCI. In *Proceedings of the SIGCHI conference on Human factors in computing systems*. 493–502.